\documentclass[journal,9pt]{IEEEtran}

\usepackage{amsmath}

\usepackage{graphicx}
\usepackage{xcolor}

\ifCLASSINFOpdf

\else

\fi

\begin{document}
%
\title{\huge{A Deep Learning Framework for Two-Dimensional, Multi-Frequency Propagation Factor Estimation}}
%
%
%

\author{\large{Sarah E. Wessinger,~\IEEEmembership{Member,~IEEE,}
        Leslie N. Smith, Jacob Gull,
        Jonathan Gehman, Zachary Beever, and Andrew J. Kammerer}

\thanks{Manuscript received April 7, 2025. This work was supported by the Office of Naval
Research via the US Naval Research Laboratory’s core funding program.}
\thanks{Sarah E. Wessinger, Andrew J. Kammerer, and Jacob Gull are with Naval Research Laboratory, Monterey, CA and Devine Consulting, Fremont, CA
USA e-mail: sarah.e.wessinger.ctr@us.navy.mil}
\thanks{Leslie N. Smith is with the US Naval Research Laboratory, Washington, DC}
\thanks{Jonathan Gehman, Zachary Beever are with Johns Hopkins University Applied Physics Laboratory, Baltimore MD}}

%
%

\markboth{IEEE Transactions on Antennas and Propagation}%
{Shell \MakeLowercase{\textit{et al.}}: Pattern Propagation Factor Prediction}
%



\maketitle

\begin{abstract}
Accurately estimating propagation factor over multiple frequencies within the marine atmospheric boundary layer is crucial for the effective deployment of radar technologies.  Traditional parabolic equation simulations, while effective, can be computationally expensive and time-intensive, limiting their practical application. This communication explores a novel approach using deep neural networks to estimate the pattern propagation factor, a critical parameter for characterizing environmental impacts on signal propagation. Image-to-image translation generators designed to ingest modified refractivity data and generate predictions of pattern propagation factors over the same domain were developed. Findings demonstrate that deep neural networks can be trained to analyze multiple frequencies and reasonably predict the pattern propagation factor, offering an alternative to traditional methods. 
\end{abstract}


%
\IEEEpeerreviewmaketitle

\section{Introduction}
\label{sec:background}
\subsection{Refractive Environment}
%
%
%
%

\IEEEPARstart{T}{he} performance of radio frequency (RF) technologies is highly susceptible to environmental factors that influence signal propagation. Fluctuations in atmospheric conditions, particularly humidity, can significantly impact signal range and quality. This variability poses challenges across diverse applications, including communication and radar technologies which are critical to national defense as well as everyday RF devices like cell phones and data-relay networks. Therefore, understanding and mitigating these environmental effects is crucial for optimizing the reliability and performance of RF systems.

The power at the end of an RF path is given by the one-way radio transmission equation \cite{friis1946note, freehafer1951theory}:
\begin{equation}
\label{eqn:transmission}
    {P_r} ={P_t} G_t G_r \left( \frac{F}{2k_0 R} \right)^2 
\end{equation}
where $P_r$ is power at a distance (slant range) \textit{R}, \textit{P\textsubscript{t} }is transmitted power, \textit{G\textsubscript{t}} is the transmitter peak power gain, \textit{G\textsubscript{r}}is the receiver peak power gain, \textit{k}\textsubscript{0} is the RF wavenumber, and $F$ is the one-way pattern propagation factor. The only variable in Equation \ref{eqn:transmission} not determined by the radar system setup is $F$ as it encapsulates all environmental impacts on signal propagation.  More specifically, $F$ is a measure of gain due to the environment. For practical applications, $F$ can also be expressed in decibels ($F_{dB}$). It is sometimes convenient to work with attenuation instead of gain using propagation loss in decibels (\textit{P\textsubscript{L}}):
\begin{equation}
    P_L = 20 \log \left( 2k_0 R \right) - 20 \log \left| F \right|
\end{equation}
\begin{equation}
\label{eqn:conv}
    F_{\text{dB}} = 10 \log \left| F \right|
\end{equation}

Accurately estimating $F$ (or, equivalently,\textit{ P\textsubscript{L}}) is imperative for quantifying environmental attenuation on RF signals. Current methods rely on parabolic equation simulations, such as the advanced propagation model (APM; \cite{barrios1994terrain}), the variable terrain radio parabolic equation (VTRPE; \cite{ryan1991user}), and the tropospheric electromagnetic parabolic equation routine (TEMPER; \cite{rouseff1992simulated}). These simulations require inputs of modified refractivity ($M$) which is calculated from atmospheric refractivity  using temperature in Kelvin ($T$), partial vapor pressure in millibar ($e$), and pressure in millibar ($p$) and adjusted to account for Earth’s curvature \cite{Bean1968radio}:
\begin{equation}
    M(z) = \left( \frac{77.6 \, p}{T} + \frac{373,256 \, e}{T^2} \right) + \frac{z}{R_e} \times 10^6
\end{equation}
where $z$ is altitude and $R_e$ is Earth’s radius. This study explores using deep neural networks (DNNs) to predict $F$ based on $M$.  DNNs excel at approximating complex nonlinear relationships by performing regressions on extensive datasets. While computationally intensive to train, once trained, DNNs offer significant advantages in terms of speed and efficiency.

DNNs and other deep learning techniques have been previously applied to RF technologies, particularly for communication applications \cite{zhang2019path,Gunduz2019machine,yang2019deep,bal2022regression,bakirtzis2022deepray,huang2025generalizable}. Notably, \cite{bal2022regression} demonstrated the potential of DNNs in estimating the path loss exponent for wireless communication, \cite{bakirtzis2022deepray} used CNNs to estimate the received signal strength of communication devices from images of physical indoor environments and demonstrated quicker computation times compared to traditional methods, and \cite{huang2025generalizable} used a CNN to solve the parabolic wave equations to estimate received signal strength over irregular terrain. However, to the authors' knowledge, no prior studies have applied DNNs to estimate $F$, $P_L$, or $F_{dB}$ over the entire domain of range and altitude.

DNNs have been used within radar applications for both forward and inverse problems \cite{wang2019study,zhu2018evaporation,pastore2022refractivity,liao2023comparison, shu2023path}. Forward problems estimate the refractive environment using environmental data, while inverse problems estimate environmental conditions or perform object detection using the refractive environment. Estimating evaporation duct height (EDH) is a popular multivariate forward problem within the refractive environment, with prior studies employing DNNs \cite{zhu2018evaporation, liao2023comparison, shu2023path}. Evaporation ducts, caused by steep near-surface vertical humidity gradients, are of interest because they trap RF waves and extend signal ranges at altitudes below the EDH. Because the layer of air where evaporation ducts form, within $\sim$30 m of the ocean surface, is often turbulent there is a high degree of self-similarity among evaporation ducts within \textit{M}(\textit{z}) profiles. This makes the estimation of EDH a generally easier problem for DNNs than learning RF signal propagation behavior based on any \textit{M}(\textit{z}). This study focuses on a forward problem where numerical weather prediction data is used to estimate the refractive environment through \textit{F}or \textit{P\textsubscript{L}}.

Characterizing RF signal propagation is especially challenging in the marine atmospheric boundary layer due to constantly fluctuating thermodynamic variables. RF technologies operating in this environment commonly use S- and X-band frequencies; thus, accurately estimating $F$ across both bands is essential. Prior studies using DNNs for multi-frequency RF problems commonly train separate models for each frequency \cite{shu2023path, popoola2019determination}. In contrast, this study trains a single DNN on a dataset comprised of two frequencies and then compares the two-frequency predictions to the individual frequency predictions. Additionally, prior research has primarily focused on attenuation as a function of only range or altitude but not both simultaneously \cite{shu2023path,nguyen2021deep, moraitis2021performance, xu2021deep}. To the authors’ knowledge, estimating \textit{F} simultaneously over both range and altitude using DNNs has not yet been attempted.

This study makes three novel contributions: (1) It demonstrates the capability of a trained DNN to accurately estimate the pattern propagation factor, (2) it achieves this estimation over both altitude and range, and (3) it introduces a unified DNN framework capable of accounting for multiple frequencies.
\label{sec:background}
\subsection{Image-to-Image Translation}
\label{sec.i2i}
Image-to-image translation is a powerful machine learning technique that enables the transformation of an image from one domain, characterized by a specific set of attributes, to another domain with new attributes, while preserving the essential content of the original image \cite{isola2017image,zhu2017unpaired}. This process involves learning the intricate mapping between the two domains, allowing for a wide range of applications in image processing, including image generation, style transfer, and segmentation \cite{pang2021image}. In this context, the \textit{M }domain (i.e., modified refractivity) is the input “image” and the $F $ domain is the desired output “image”.

In recent years, deep learning frameworks have revolutionized the field of image-to-image translation, with autoencoders, convolutional neural networks (CNNs), and generative adversarial networks (GANs) achieving remarkable success. GANs, in particular, have emerged as a popular choice for image-to-image translation tasks \cite{pang2021image,alotaibi2020deep}, due to their ability to learn complex mappings between domains. A typical GAN architecture for image-to-image translation consists of two primary components: a generator network and a discriminator network. The generator network produces an image in the output domain, while the discriminator network evaluates the generated images, providing feedback to the generator to improve its performance.

\subsection{Neural Network-Based Emulator Architecture}
\label{sec:arch}
Initially, GAN architecture was implemented for image-to-image translation, as it is a widely accepted architecture for this task. The discriminator in a GAN plays a crucial role in guiding the generator by assessing the realism of the generated images. Through adversarial training, the discriminator enforces realism by distinguishing fine details, textures, and patterns that align with the desired target distribution. This feedback loop enables the generator to iteratively refine its outputs, ensuring they match the statistical properties of the real data. However, in this study, supervised training with explicitly desired outputs renders the adversarial training unnecessary and only the generator was needed.

U-Net architectures \cite{ronneberger2015u} are commonly employed as the generator in GANs \cite{ronneberger2015u}. Their architecture is characterized by a contracting and an expanding block, connected by skip connections and an inner module. Skip connections enable the model to detect features at multiple resolutions, capturing both global structural relationships and pixel-wise details. This design allows the U-Net to effectively learn the complex mappings between domains, making it an ideal choice for image-to-image translation tasks.

This study evaluated the merits of implementing a variety of UNet-based generator architectures. The following lists the models tested in order of size and complexity:
\begin{itemize}
	\item Two versions of U-Net \cite{ronneberger2015u}
	\item Recursive Neural Network (RNN)
	\item Efficient U-Net \cite{tan2019efficientnet}
	\item Dilated U-Net \cite{wang2020u}
	\item UVCGan v1 and v2 (Visual Transformer at bottleneck) \cite{torbunov2023rethinking, torbunov2023uvcgan}
	\item UCTransNet (transformer instead of skip connections) \cite{wang2022uctransnet}
	\item A novel hybrid (combines UVCGan and UCTransNet) 
\end{itemize}
We observed significant improvements in performance from the first U-Nets test to the larger and more complex models. However, our evaluations demonstrated that architectural innovations alone were approaching the limits of significant performance gains. While initial gains underscore the importance of exploring alternative architectures, the final two models only provided minor performance improvements but with an increase in computational cost. Consequently, UVCGANv2 is the default model used in this study.  The UVCGANv2 architecture, henceforth referred to as "RefractNet", is illustrated in \cite{torbunov2023uvcgan}. RefractNet's training hyperparameters are listed in Table \ref{table_HP} for the sake of reproducibility. The Bayesian optimization method Optuna was used to tune these parameters.

\begin{table}[!t]
\renewcommand{\arraystretch}{1.3}
\caption{RefractNet Training Hyper-parameters}
\label{table_HP}
\centering
\begin{tabular}{|c|c|}
\hline
Hyper-parameter & Value\\
\hline
\hline
Batch size  & 140\\
\hline
Num Epochs  & 100 or 200\\
\hline
Learning rate (LR)  & 0.0001\\
\hline
LR schedule  & 1cycle\\
\hline
Dropout  & 0.02\\
\hline
Num uvcgan2 blocks & 12\\
\hline
Num uvcgan features & 384\\
\hline
\end{tabular}
\end{table}


\section{Methods}
\subsection{Data}
$M(z)$ used within this study is from the Naval Research Laboratory’s numerical weather prediction system, the Coupled Ocean Atmosphere Mesoscale Prediction System (COAMPS®). COAMPS® forecasts were run for 7 arbitrary oceanic locations across the globe to remove local geographic dependence on these results. 

The physics-based parabolic simulation used to calculate the refractive domain of $F$ from $M(z)$ is TEMPER version 3.2.3. TEMPER was run for two frequencies, 3 and 10 GHz (S- and X-Band, respectively). TEMPER's mesh size to calculate F is approximately half the frequency wavelength, assuming a smooth sea surface. Rough ocean surfaces could be an added complexity at a later time; the point of this study is an initial proof-of-concept focusing on refractivity, \textit{M}(\textit{z}). The antenna height was set to 20 m for each simulation. 

TEMPER simulations for S- and X-band have varying domain output altitudes based on mean sea level. All S-Band runs range in altitude from MSL (0m) to 300m, whereas X-band runs extend from the MSL (0m) to 30m, both at a 0.2 m vertical resolution. All S-Band runs have a horizontal range from 0 to 210 km at a 0.2 km resolution, whereas X-band runs extend from 0 to 100 km, at a 0.1 km resolution.

\begin{figure}[!t]
\centering
\includegraphics[width=3.5in]{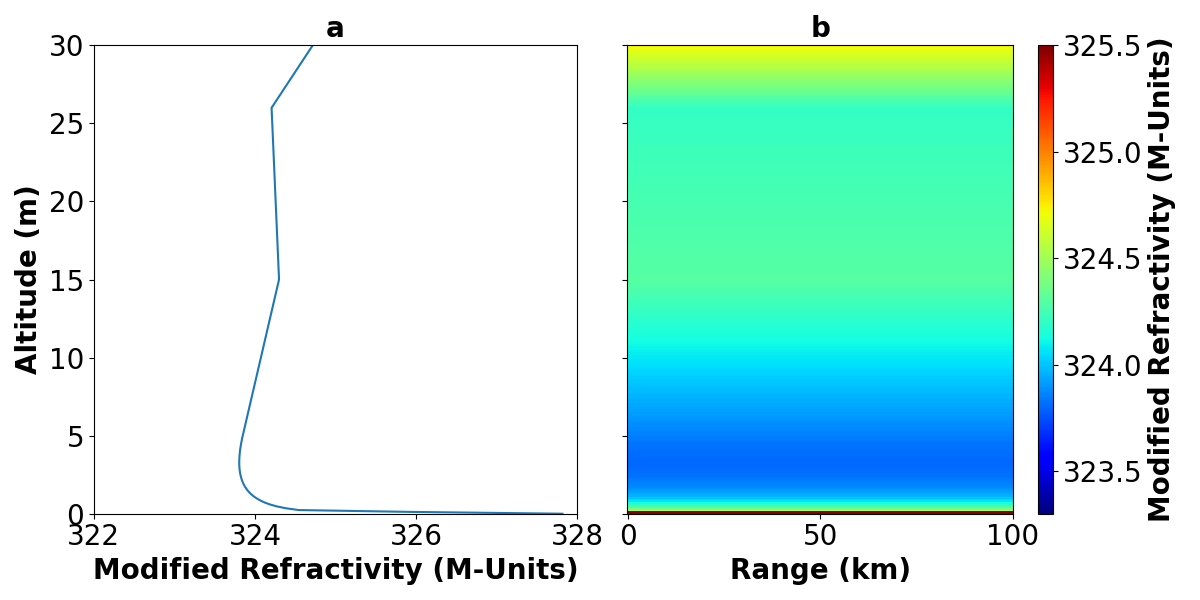}
\caption{(a) $M(z)$ used to create the (b) range-homogeneous $M$ domain input for RefractNet.}
\label{fig_modrefexample}
\end{figure}

This study used 70,322 total cases, comprised of an $M(z)$ input and a $F$ domain output to train RefractNet, and 15,070 total cases were used to evaluate its performance. Half of all cases, for both test and train datasets, are X-Band and half are S-Band frequency. We expect this number of training samples to be sufficient to train our network because several well-known image benchmark datasets contain on the order of 50,000 training samples (i.e., MNIST, Cifar-10/100). Note, $M$ is assumed homogeneous over range and time for this study (Figure \ref{fig_modrefexample}). 

\subsection{Metrics}
\label{sec.metrics}

In addition to visual inspection of individual examples, this study leveraged a combination of three key performance metrics: Mean Squared Error (MSE), Structural Similarity (SSIM), and Fréchet Inception Distance (FID).  Mean Squared Error (MSE) is a widely used metric that calculates the average squared difference between the pixel values of real and generated images, penalizing larger errors more heavily. However, as a pixel-wise measure, MSE is limited in capturing larger-scale structural trends, which are critical in evaluating the accuracy of complex images. In contrast, Structural Similarity (SSIM) \cite{wang2004image} assesses error while considering structural information and broader trends, yielding a score between 0 (no similarity) and 1 (perfect similarity). Utilizing convolutional properties, SSIM effectively identifies low-level structures within images, which is critical for evaluating the accuracy of the main structure. Fréchet Inception Distance (FID) \cite{yu2021frechet} is the most common metric used to evaluate GAN results as it compares features extracted from real and generated images using a pre-trained network, with lower scores indicating better fidelity to real images. SSIM and FID are particularly useful in evaluating the performance of the emulator as they provide a more comprehensive and nuanced understanding of the output quality.

Each metric can be computed for individual samples, batches, frequency, and depth categories, enabling clustering analyses to identify classes of inputs where the model performs poorly. This is particularly useful in identifying areas where the model requires improvement and guides the approach used to optimize the emulator's performance. As such, during RefractNet's development, these metrics guided efforts to improve the model's accuracy through various methods, including data augmentation, loss function optimization, hyper-parameter tuning, and experimentation with different U-Net architectures. We tested several loss function variations and settled on an equal combination of MSE and SSIM for our loss function.

\section{Experiments}
\label{sec:exp}

\subsection{Setup}
\label{sec:setup}
Eight experiments were conducted during this study (Table \ref{table_experiments}). Input/output domains having 256 x 256 pixels is optimal for a balance between performance and speed. To meet this requirement, $M(z)$ is interpolated to have 256 equally spaced points over the evaluated altitudes (i.e., 0-30m or 0-300m) and repeated 256 times over range (Figure \ref{fig_modrefexample}). The corresponding domains of $F$ were interpolated to have the same points over altitude as the $M$ domains (256 x 256) and equally spaced points over range between 0-100 km so both frequencies are evaluated over the same physical space. Input and output domains were scaled to have values between 0 and 1 due to batch normalization layers within the architecture. As such, all data points for $M$ domain images and $F$ domain images were normalized based on altitude and gain variable following the normalization schemes in Table \ref{table_norm} based on Equation \ref{normalizationScheme} where $V$ is the variable being normalized ($M$, $F$, etc.): 
\begin{equation}
    V(z,x) = \frac{V(z,x)-V_{min}}{V_{max}-V_{min}}
\label{normalizationScheme}
\end{equation}

 
\begin{table}[!t]
\renewcommand{\arraystretch}{1.3}
\caption{Combinations of variables used to train the RefractNet for each experiment}
\label{table_experiments}
\centering
\begin{tabular}{|c||c|}
\hline
Experiment Number & Variable, Altitude, Frequencies\\
\hline
Experiment 1 & $F$, 30 m, X + S\\
\hline
Experiment 2 & $F$, 30 m, X\\
\hline
Experiment 3 & $F$, 30 m, S\\
\hline
Experiment 4 & $F_{dB}$, 30 m, X + S\\
\hline
Experiment 5 & $F_{dB}$, 30 m, X\\
\hline
Experiment 6 & $F_{dB}$, 30 m, S\\
\hline
Experiment 7 & $F$, 300 m, S\\
\hline
Experiment 8 & $F_{dB}$, 300 m, S\\
\hline
\end{tabular}
\end{table}


\begin{table}[!t]
\renewcommand{\arraystretch}{1.3}
\caption{Normalization Schemes Used Based on Altitude and Gain Variable}
\label{table_norm}
\centering
\begin{tabular}{|c||c||c|}
\hline
Altitude, Variable & $M(z)$ Norm. & $F(z,x)$ Norm.\\
\hline
30 m, $F$ & \(M = \frac{M-288}{181}\) & \(F = \frac{F}{16.45}\)\\
\hline
300 m, $F$ & \(M = \frac{M-282}{187}\) & \(F = \frac{F}{16.45}\)\\
\hline
30 m, $F_{dB}$ & \(M = \frac{M-288}{181}\) & \(F_{dB} = \frac{F_{dB}+90.01}{-102.17}\)\\
\hline
300 m, $F_{dB}$ & \(M = \frac{M-282}{187}\) & \(F_{dB} = \frac{F_{dB}+90.01}{-102.17}\)\\
\hline
\end{tabular}
\end{table}

In addition, single-frequency experiments have half the number of training samples as the dual-frequency experiments, as previously described.  Thus, RefractNet was trained for twice as many epochs for single-frequency experiments in order to minimize any training advantage the dual-frequency experiments might gain by having twice as many training samples. 

\subsection{Results}

\begin{figure*}
\centering
\includegraphics[width=.65\textwidth]{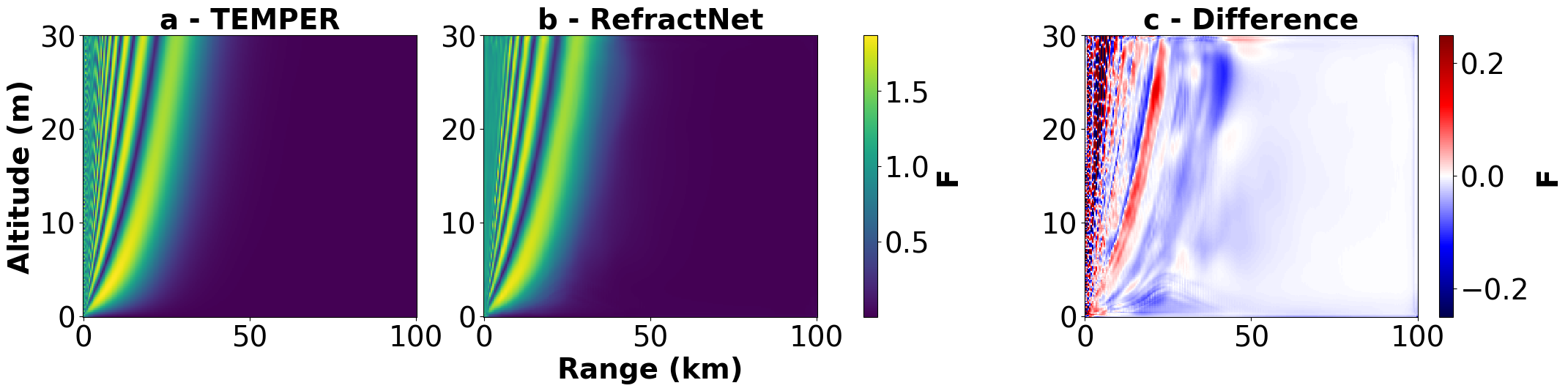}
\caption{$F$ domain calculated by (a) TEMPER, (b) RefractNet, and (c) the difference between these domains, TEMPER-RefractNet, from a case in Experiment 1 where the evaluation metrics are: MSE = 1.086x10\textsuperscript{-5}, FID = 0.0016, and SSIM = 0.930.}
\label{Picture_domainEx}
\end{figure*}

\begin{figure}[!t]
\centering
\includegraphics[width=3.55in]{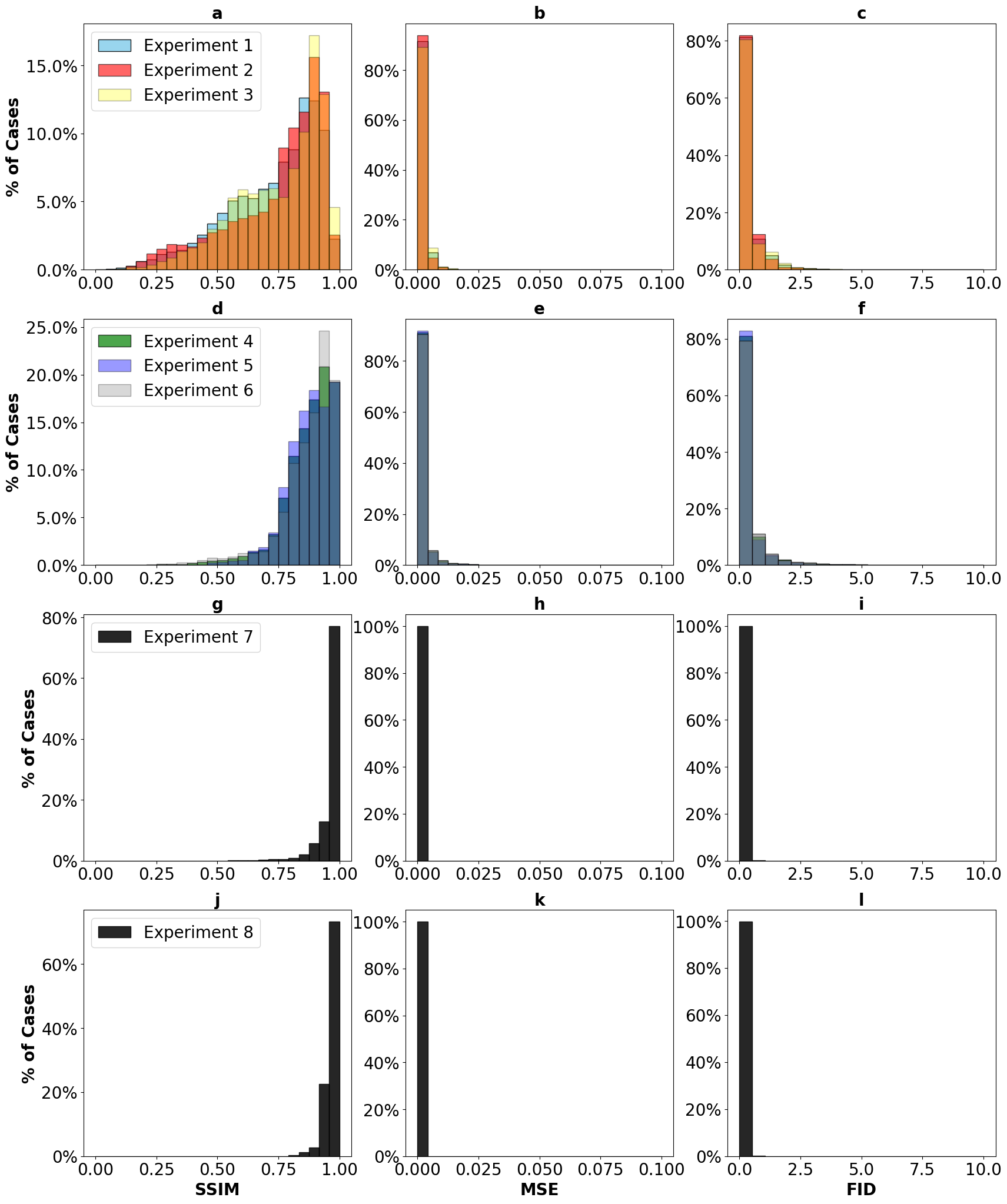}
\caption{Distribution of evaluation metrics for SSIM (a, d, g, j), MSE (b, e, h, k), and FID (c, f, i, l) for Experiments 1-3 (a-c), 4-6 (d-f), 7 (g-i), and 8 (j-l).}
\label{results_dist}
\end{figure}

\begin{figure*}
\centering
\includegraphics[width=.65\textwidth]{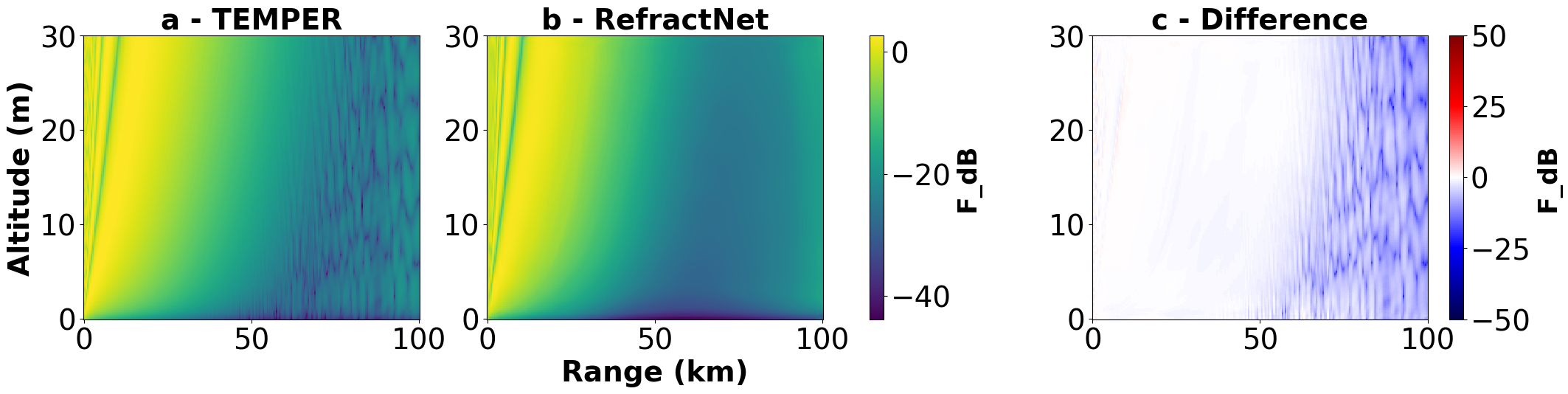}
\caption{$F_{dB}$ domain calculated by (a) TEMPER, (b) RefractNet, and (c) the difference, TEMPER-RefractNet, between these domains for a case from Experiment 6 where the evaluation metrics are: MSE = 0.0020, FID = 0.515, and SSIM = 0.758.}
\label{BadEx}
\end{figure*}

Each evaluation metric (MSE, FID, SSIM) is calculated between the  RefractNet-predicted $F$ domain image and TEMPER's normalized $F$ domain image for each test case in each experiment (Figure \ref{Picture_domainEx}), thus these values are dimensionless. Distributions of evaluation metrics, presented in Figure \ref{results_dist}, demonstrate that RefractNet performs well across a variety of scenarios.  The majority of cases exhibit high SSIM values (indicating strong structural similarity between predictions and TEMPER) and low FID and MSE values (demonstrating low dissimilarity and error). Figure \ref{BadEx} illustrates a case that performed poorly due to having ducting (trapping) features at longer ranges, which are indicative of signal interference due to environmental reflection and/or refraction. 

To rigorously assess RefractNet's capacity to handle multiple frequencies, performance of single-frequency experiments against their dual-frequency counterparts was statistically compared. Specifically, two-sample t-tests were performed on the MSE, SSIM, and FID populations for the following experiment pairs, ensuring consistent altitude and gain variables: Experiments 1 \& 2, Experiments 1 \& 3, Experiments 4 \& 5, and Experiments 4 \& 6. P-values from Experiments 1 \& 2, Experiments 1 \& 3, and Experiments 4 \& 5 (FID and MSE only) of the t-tests were $<$0.05, indicating these null hypotheses (e.g., Average MSE Experiment 1 = Average MSE Experiment 3) are rejected, and the average values (Table \ref{table_avgs}) of these evaluation metrics are different in a statistically significant sense. Interestingly, the null hypotheses in the case of Experiments 4 \& 6 are accepted, suggesting these average values are not statistically significantly different.

\begin{table}[!t]
\renewcommand{\arraystretch}{1.3}
\caption{Average values of evaluation metrics for each experiment.}
\label{table_avgs}
\centering
\begin{tabular}{|c||c||c||c|}
\hline
Experiment Number & MSE & FID & SSIM\\
\hline
Experiment 1 & 0.0010 & 0.259 & 0.724\\
\hline
Experiment 2 & 0.0009 & 0.230 & 0.742\\
\hline
Experiment 3 & 0.0011 & 0.284 & 0.752\\
\hline
Experiment 4 & 0.0017 & 0.439 & 0.871\\
\hline
Experiment 5 & 0.0016 & 0.397 &  0.873\\
\hline
Experiment 6 & 0.0017 & 0.442 & 0.870\\
\hline
Experiment 7 &  0.0001 & 0.010 & 0.965\\
\hline
Experiment 8 & 0.0001 & 0.025 & 0.971\\
\hline
\end{tabular}
\end{table}

Interestingly, X-band-only experiments (Exp. 2 \& 5) consistently outperformed both S-band-only (Exp. 3 \& 6) and dual-frequency experiments (Exp. 1 \& 4) across all metrics (Table \ref{table_avgs}), albeit not by much. This highlights an important consideration: RF signal propagation becomes increasingly sensitive to atmospheric refractive structures at higher frequencies \cite{lentini2015global}. Consequently, the $F$ domain exhibits more distinct and repetitive patterns at higher frequencies, potentially contributing to more robust and effective DNN training. 

\begin{table}[!t]
\caption{Average values of evaluation metrics for each frequency in dual frequency experiments.}
\label{ind_avgs}
\centering
\begin{tabular}{|c||c||c||c|}
\hline
Experiment Number, Frequency & MSE & FID & SSIM\\
\hline
Experiment 1, X & 0.0010 & 0.235 & 0.706\\
\hline
Experiment 1, S & 0.0011 & 0.283 & 0.743\\
\hline
Experiment 4, X & 0.0016 & 0.408 & 0.873\\
\hline
Experiment 4, S & 0.0018 & 0.469 & 0.869\\
\hline
\end{tabular}
\end{table}

\begin{figure*}
\centering
\includegraphics[width=.65\textwidth]{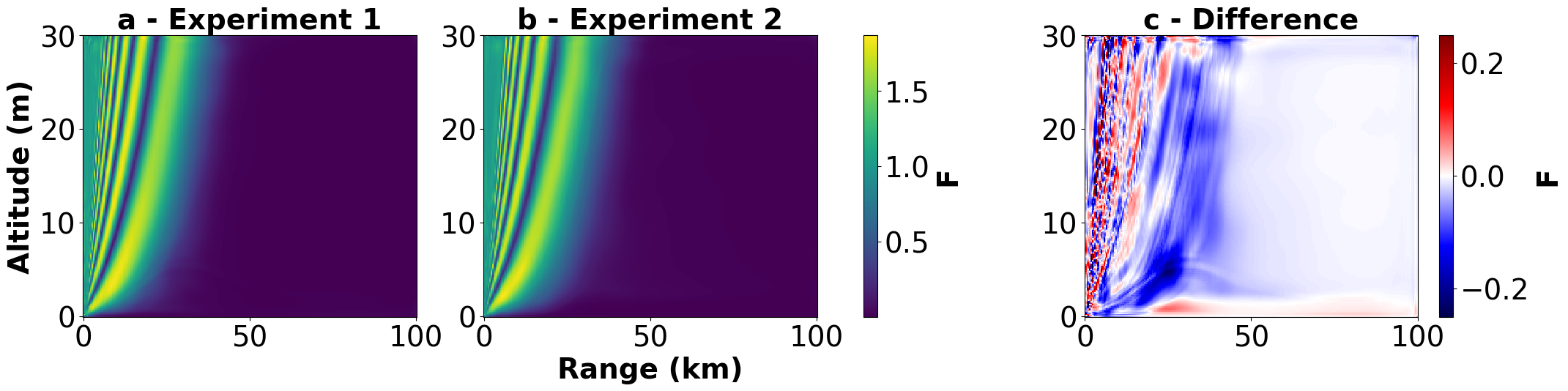}
\caption{$F$ domain for the same case in both (a) Experiment 1 with metrics of MSE = 1.086x10\textsuperscript{-5}, FID = 0.0016, and SSIM = 0.930, and (b) Experiment 2 of MSE = 8.035x10\textsuperscript{-6}, FID = 0.0012, and SSIM = 0.843 and (c) the difference between these domains.}
\label{domainFrequencyComp}
\end{figure*}

Despite evaluation metrics performing better for X-band-only training, dual-frequency-trained experiments yield similar patterns within the $F$/$F_{dB}$ domains (Figure \ref{domainFrequencyComp}), and similar average evaluation metric values for each individual frequency in dual-frequency experiments (Table \ref{ind_avgs}) compared to single-frequency-trained experiments. Thus, this DNN framework is able to account for two frequencies. 

To further investigate RefractNet's performance, predictions of $F$ versus $F_{dB}$ domains are evaluated. RefractNet generally performed better when predicting $F$ as shown by average MSEs and FIDs. This difference likely arises from the logarithmic conversion from $F$ to $F_{dB}$ (Equation \ref{eqn:conv}), which amplifies the influence of refractive structures present in the refractive domains (as evidenced by the larger average SSIM values), especially at longer ranges (Figure \ref{BadEx}), during training. This suggests DNNs should be trained to estimate $F$, with subsequent calculation of desired gain or attenuation variables (e.g., $F_{dB}$, $P_L$) derived from the predicted $F$ values.

\begin{figure*}
\centering
\includegraphics[width=.65\textwidth]{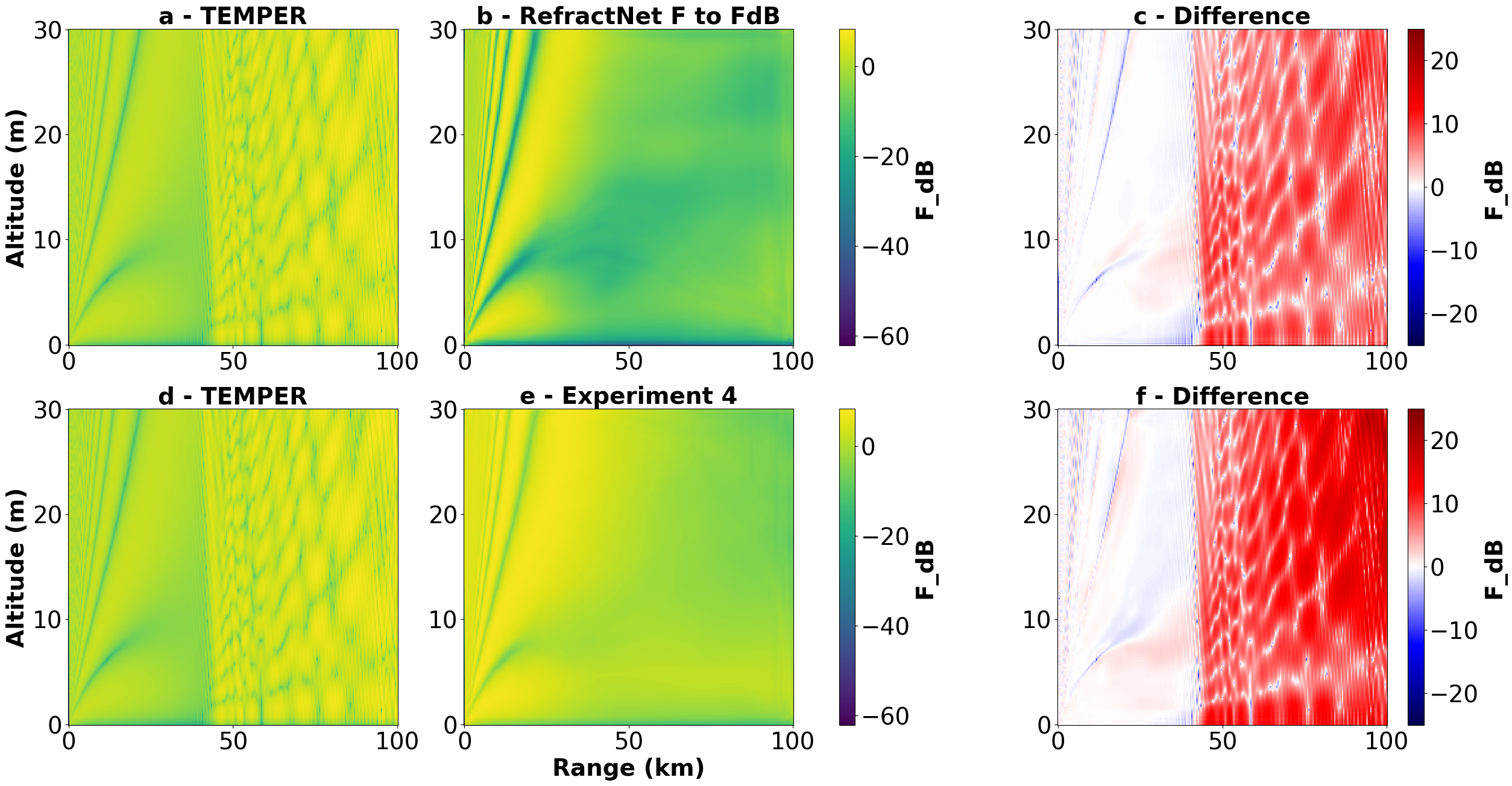}
\caption{$F_{dB}$ domain (a,d) from TEMPER, (b) calculated from Experiment 1 $F$ (c) the difference between a and b, (e) from Experiment 4, and (f) the difference between d and e.}
\label{FComp}
\end{figure*}

\begin{figure*}
\centering
\includegraphics[width=.65\textwidth]{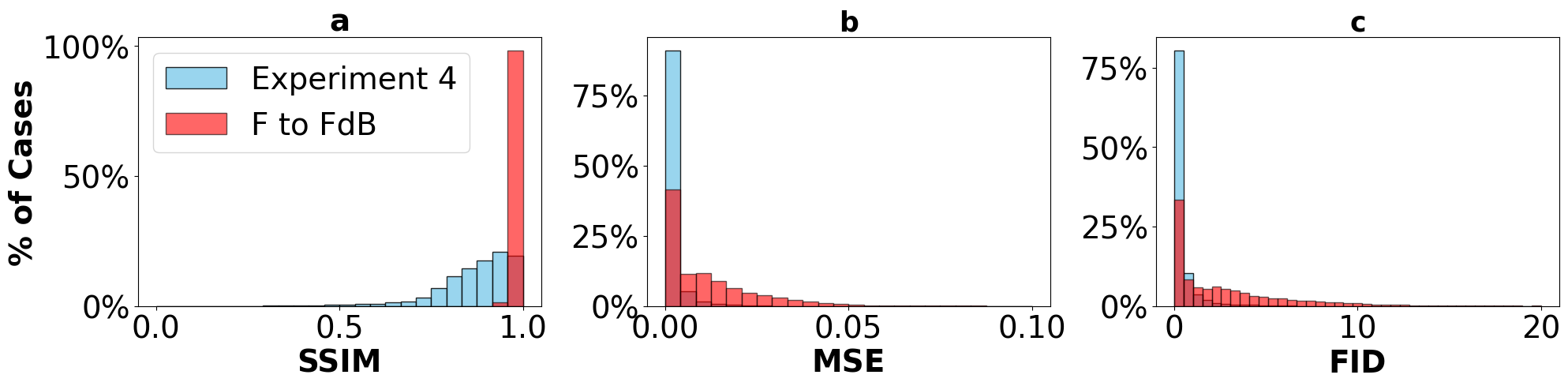}
\caption{Distributions of (a) SSIM, (b) MSE, and (c) FID for Experiment 4 compared to the calculated $F_{dB}$ from Experiment 1 $F$.}
\label{FComp2}
\end{figure*}

To explore this suggestion, RefractNet's predicted $F$ domain images from Experiment 1 (i.e., trained on $F$) are converted to $F_{dB}$ (Equation \ref{eqn:conv}), and compared to TEMPER's $F_{dB}$ domain images using the evaluation metrics. These evaluation metrics are compared to Experiment 4 (Figure \ref{FComp}), which was trained on $F_{dB}$ and compared RefractNet's predicted $F_{dB}$ domain images to TEMPER's $F_{dB}$ images. Results show images of $F_{dB}$ converted from RefractNet-generated $F$ generally have greater structural similarity to TEMPER's predicted $F_{dB}$ domain images than Experiment 4 but greater errors in estimated values (Figure \ref{FComp2}). 

\begin{figure*}
\centering
\includegraphics[width=.65\textwidth]{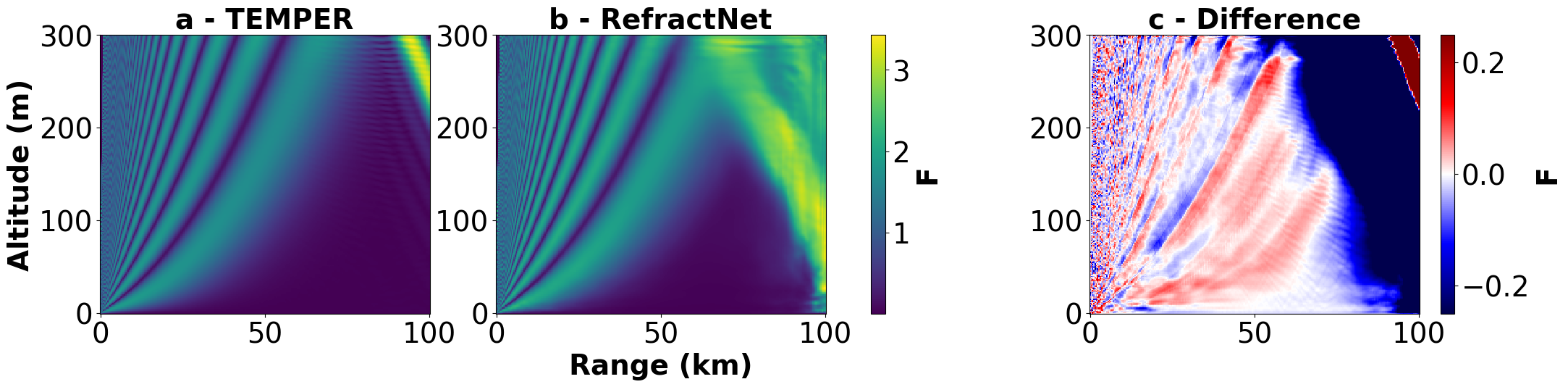}
\caption{$F$ domain calculated by (a) TEMPER, (b) RefractNet, and (c) the difference between these domains for a case from Experiment 7 where the evaluation metrics are: MSE = 0.0021, FID = 0.505, and SSIM = 0.749.}
\label{F300}
\end{figure*}

Comparing the performance of all S-Band experiments, higher-altitude experiments consistently outperformed lower-altitude. This finding aligns with expectations, as higher-altitude scenarios contain more data points within the radar horizon and fully resolve patterns of constructive interference that occur over 30 m (Figure \ref{F300}). Conversely, lower altitude scenarios contain more pixels outside of the radar horizon and don't fully resolve propagation structures occurring over 30 m (Figure \ref{BadEx}). Therefore, future studies developing DNNs for radar applications aimed at identifying specific propagation features should carefully consider over what altitudes and ranges to evaluate and potentially even exclude data points within the radar horizon to enhance model training and accuracy.

\section{Conclusion}
This study introduced RefractNet, an image-to-image DNN generator designed for efficient simulation of the propagation factor, over both range and altitude, while accounting for multiple frequencies. RefractNet exhibited a slightly superior performance for X-band frequencies compared to S-band. From a practical standpoint, RefractNet demonstrated the capacity to handle both frequencies simultaneously. These results hold significant promise for the future of trained DNNs as a viable method to estimate propagation factor, but also illustrate the sensitivity of predicting propagation factor using DNNs. Results of this study varied considerably between training on different domain altitudes, $F$ or $F_{dB}$, single or multiple frequencies, and even training on $F$ and assessing performance using $F_{dB}$, thus, future work estimating the propagation factor using DNNs needs to carefully consider how training is approached. 



%



\section*{Acknowledgment}
The authors acknowledge the support by the Office of Naval Research via the US Naval Research Laboratory's core funding program. LNS acknowledges the continued conversations and support of Dr Josette Fabre of NRL's Code 7180.
\ifCLASSOPTIONcaptionsoff
  \newpage
\fi



%
\bibliographystyle{IEEEtran}
\bibliography{scholar}
\end{document}